\pdfoutput=1

\documentclass[11pt]{article}
\usepackage{hyperref}
\usepackage{marvosym}

\usepackage{changepage} 
\usepackage{placeins}
\usepackage{marvosym} 

\usepackage[]{ACL2023}

\usepackage{times}
\usepackage{xspace}
\usepackage{xfp}
\usepackage{latexsym}
\usepackage{xcolor}
\usepackage{soul}
\usepackage[table]{xcolor}

\definecolor{FindingsLongColor}{HTML}{FDBE85}  
\definecolor{FindingsShortColor}{HTML}{8DA0CB} 
\definecolor{MainLongColor}{HTML}{FC8D62}      
\definecolor{MainShortColor}{HTML}{A8E6A3}

\definecolor{Introduction}{HTML}{5A049C}
\definecolor{QuestionBlue}{HTML}{0060A5}
\definecolor{AdditionalContext}{HTML}{882200}
\definecolor{SectionNames}{HTML}{5BA50C}
\definecolor{OutputStructure}{HTML}{012321}

\usepackage{booktabs}
\usepackage{tabularx}
\usepackage{pifont}
\usepackage[T1]{fontenc}
\usepackage[utf8]{inputenc}
\usepackage{microtype}

\usepackage{inconsolata}

\usepackage{url}
\usepackage[T1]{fontenc}
\usepackage{multirow}
\usepackage{listings}
\usepackage{graphicx}

\usepackage{amsmath}

\graphicspath{ {./figures/} }

\newcommand{\numpapers}{1975\xspace}
\newcommand{\numpaperseval}{186\xspace}

\newcommand{\numquestionsint}{19}

\newcommand\blfootnote[1]{%
  \begingroup
  \renewcommand\thefootnote{}%
  \footnote{#1}%
  \addtocounter{footnote}{-1}%
  \endgroup
}

\title{ConfReady: A RAG based Assistant and Dataset for Conference Checklist Responses}

\author{
Michael Galarnyk\textsuperscript{1*\,\Letter}\quad
Rutwik Routu\textsuperscript{2*\,\textdagger}\quad
Vidhyakshaya Kannan\textsuperscript{3*\,\textdagger}\quad
Kosha Bheda\textsuperscript{1}\\
\textbf{Prasun Banerjee}\textsuperscript{1}\quad
\textbf{Agam Shah}\textsuperscript{1}\quad
\textbf{Sudheer Chava}\textsuperscript{1}\\[0.5em]
\textsuperscript{1}\,Georgia Institute of Technology\quad
\textsuperscript{2}\,Duke University\quad
\textsuperscript{3}\,Sai University\\
{\Letter\;\;Corresponding author: \href{mailto:mgalarnyk3@gatech.edu}{mgalarnyk3@gatech.edu}}
}

\begin{document}
\maketitle

\blfootnote{*\;These authors contributed equally.}
\blfootnote{\textdagger\;\;Work done as a Volunteer Research Assistant at Georgia Institute of Technology.}

\begin{abstract}
The ARR Responsible NLP Research checklist website states that the "checklist is designed to encourage best practices for responsible research, addressing issues of research ethics, societal impact and reproducibility." Answering the questions is an opportunity for authors to reflect on their work and make sure any shared scientific assets follow best practices. Ideally, considering a checklist before submission can favorably impact the writing of a research paper. However, previous research has shown that self-reported checklist responses don't always accurately represent papers. In this work, we introduce ConfReady, a retrieval-augmented generation (RAG) application that can be used to empower authors to reflect on their work and assist authors with conference checklists. To evaluate checklist assistants, we curate a dataset of \numpapers ACL checklist responses, analyze problems in human answers, and benchmark RAG and Large Language Model (LM) based systems on an evaluation subset. Our code is released under the AGPL-3.0 license on \href{https://github.com/gtfintechlab/ConfReady}{GitHub}, with \href{https://confready-docs.vercel.app}{documentation} covering the user interface and \href{https://pypi.org/project/confready/}{Python package}.
\end{abstract}

\section{Introduction} 
In order to submit a paper to conferences under the Association for Computational Linguistics like ACL, COLING, CoNLL, EMNLP, and NAACL, authors are required to submit their answers to the ARR Responsible NLP Research checklist\footnote{\url{https://aclrollingreview.org/responsibleNLPresearch/}}. The checklist was mostly developed through a combination of the NLP Reproducibility Checklist \cite{dodge-etal-2019-show}, the reproducible data checklist \cite{rogers-etal-2021-just-think}, and the NeurIPS 2021 Paper Checklist Guidelines\footnote{\url{https://neurips.cc/Conferences/2021/PaperInformation/PaperChecklist}}. The goal of this process is to address reproducibility, societal impact, and potential ethical issues. Authors are expected to discuss limitations, artifact usage, computational details, human involvement, and use of AI assistants. Starting with EMNLP 2025, checklist responses will be published as appendices alongside accepted papers\footnote{\url{https://aclrollingreview.org/responsible-nlp-checklist-appendices}}, in order to "help with transparency" and encourage authors to "think more carefully about these issues when they know their answers will be visible to the broader community."

This follows an earlier pilot at ACL 2023, where checklist responses—covering 19 questions per paper—were appended to accepted submissions. For example, question A2 asks: \textit{"Did you discuss any potential risks of your work?"} If authors respond ``yes,'' they must cite the relevant section; if ``no,'' they are expected to provide a justification.  However, prior work has noted cases of low-effort or bad-faith responses, such as identical answers across questions and falsely reporting code availability \citep{magnusson-etal-2023-reproducibility}. The ACL 2023 program chairs suggested that checklist sloppiness correlates with sloppiness elsewhere in the work and a lower acceptance rate \citep{rogers-etal-2023-report}.

\begin{figure*}[t]
     \centering\includegraphics[width=\textwidth]{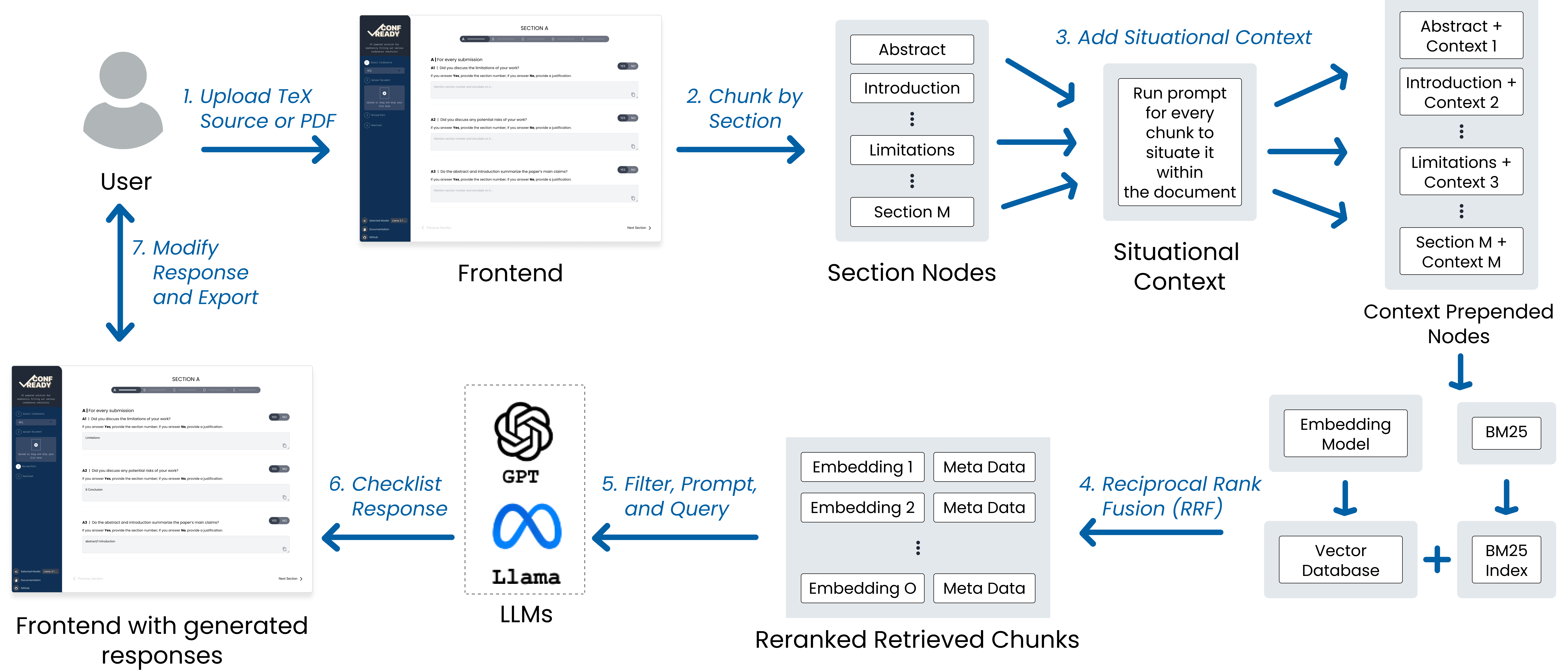}
     \caption{Users can upload either TeX source (single .tex file or a zipped folder) or a PDF to the frontend to receive an LM-generated checklist response, which can then be modified and exported.}
     \label{fig:overallApp}
 \end{figure*}

To mitigate unreliable checklist answers, conferences have explored Large Language Model (LM) based systems to assist authors \cite{goldberg2024usefulnessllmsauthorchecklist}, though these were not evaluated against human-written checklist submissions and do not reflect the full complexity of real submissions. To address this, LMs can be augmented with tools like retrieval-augmented generation (RAG), which integrates information retrieval with generative models \cite{lewis2020retrieval}. This approach improves accuracy and relevance, especially for question-answering tasks requiring up-to-date or domain-specific knowledge \cite{karpukhin2020dense}. 

In this work, we introduce \textbf{ConfReady}, a RAG tool that helps with checklist responses grounded in their paper’s TeX source or PDF\footnote{A video demonstration is available at \url{https://youtu.be/sNhpKJLfArc?si=0CMCe1nEFwFFUibw}, with documentation and a pip-installable package at \url{https://confready-docs.vercel.app}.}. To evaluate ConfReady, we compile the ConfReady dataset of real checklist submissions (see Section~\ref{sec:dataset}) and evaluate its outputs against human-written answers (see Section~\ref{sec:evaluation}). 

Our contributions are the following:
\begin{itemize}
    \item \textbf{ConfReady Tool}: An end-to-end system for generating checklist responses from TeX source (single .tex file or zipped folder) or PDF, with a user-friendly interface and pip-installable backend.
    \item \textbf{Checklist Dataset}: A structured dataset of \numpapers ACL papers with parsed checklist responses and metadata, enabling analysis and benchmarking.
    \item \textbf{Backend Evaluation}: A comparison of RAG and LM only backends on 93 ACL papers, with accuracy measured against human-provided checklist responses.
\end{itemize}

\section{ConfReady} 
\label{sec: confready}
\begin{figure*}[t]
     \centering
         \includegraphics[width=\textwidth]{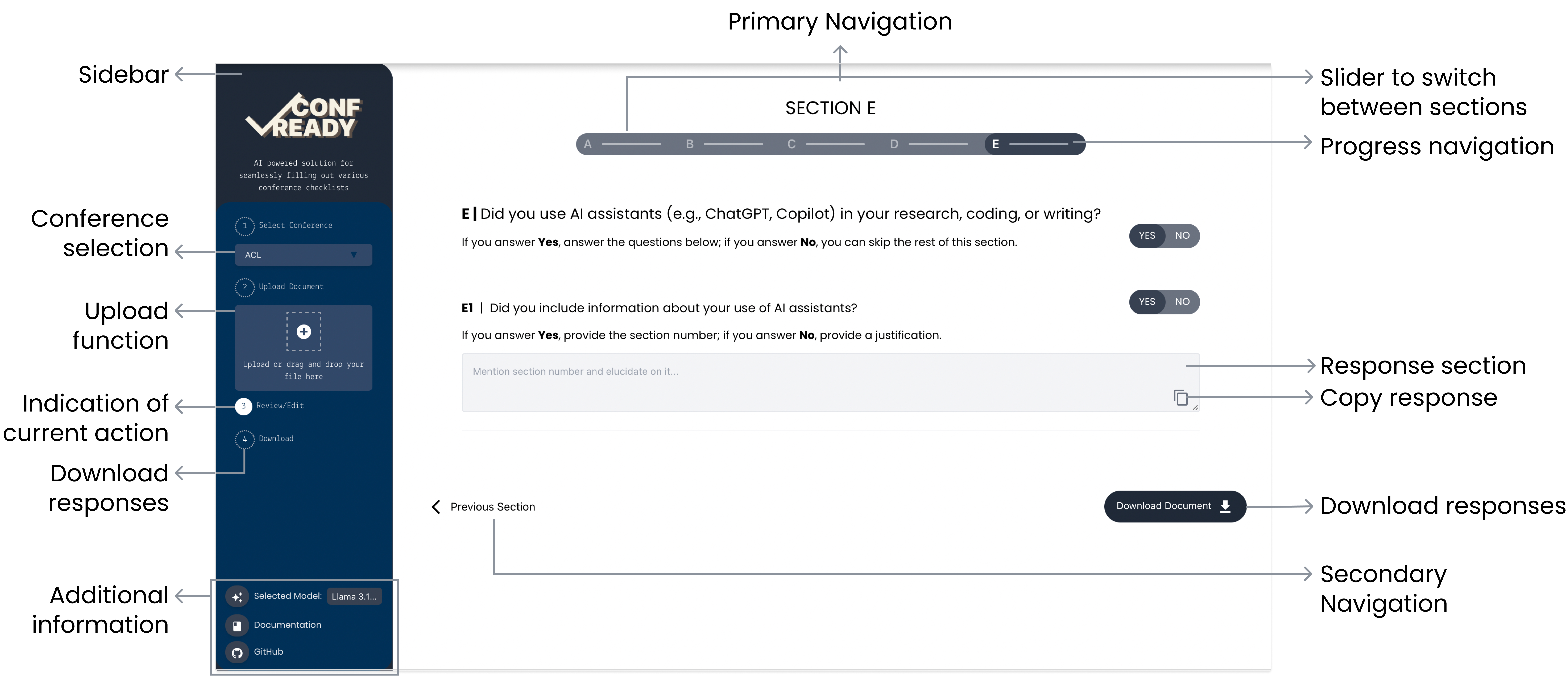}
     \caption{Features of the ConfReady user interface include: an upload function within the sidebar, primary navigation with a slider to switch between sections and progress navigation, and a generated response field with a copy function. The design rationale behind the features is listed in Appendix \ref{app:features_userinference}.}
    \label{fig:UIChecklist}
\end{figure*}

Figure~\ref{fig:overallApp} presents the ConfReady pipeline. Users start by uploading either TeX source (single .tex or zipped folder) or a PDF on the frontend. The system then processes the document through a seven-step workflow: (1) upload input, (2) chunk by section, (3) add situational context to each section, (4) perform Reciprocal Rank Fusion (RRF), (5) construct and send prompts to the LM, (6) generate checklist responses, and (7) allow users to review, edit, and export the results.
The user-facing interface, shown in Figure \ref{fig:UIChecklist}, supports this workflow with features like section-level navigation, editable response fields, and progress tracking. Appendix \ref{app:features_userinference} details the design rationale behind each interface component.

\subsection{Parsing, Chunking, and Embedding}

\paragraph{Parsing} Users can upload either TeX Source (single .tex file or zipped folder) or PDF. When TeX Source is provided, the document is parsed to remove comments and pre-abstract content. For PDFs, a simplified pipeline is applied.

\paragraph{Contextual Chunking}
ConfReady uses contextual chunks \cite{anthropic2024contextual} to reduce retrieval failure rates. 
Each chunk is annotated with metadata (e.g., section title, neighboring chunk identifiers), and a short LM-generated situational summary is prepended to the chunk before embedding using the following prompt: 

\begin{quote}
\textbf{Prompt:} Provide a concise (50–100 tokens) situational summary for this chunk, capturing its role in the larger section.
\end{quote}

\paragraph{Embeddings} The enriched chunks (metadata, situational summary, original text) are then converted into vector representations and stored locally. The embedding backend is modular and can be configured to support different models.

\subsection{Retrieval, Fusion, and Reranking}
\paragraph{Retrieval}

Our retrieval workflow is designed to combine the strengths of dense vector embeddings and sparse lexical search. 

\paragraph{Dense Retrieval with Embeddings} Each chunk is embedded using OpenAI’s \texttt{text-embedding-3-large}. Queries are also embedded and compared against the document embeddings using cosine similarity.

\paragraph{Sparse Retrieval with BM25} In parallel, we build a lexical index over all contextualized chunks using the BM25 algorithm. This sparse retrieval step is especially effective at capturing exact matches, uncommon terminology, and keyword-based relevance that dense models may miss.

\paragraph{Score Fusion (Vector + BM25)} To balance semantic and lexical relevance, we perform RRF on the top results from both the dense and sparse retrieval components.

\paragraph{LM-Based Reranking} The fused results are passed through a reranking module, which uses an LM to evaluate the relevance of each chunk in the context of the original query. The top-k chunks from this reranking step are selected as final inputs for generation.

\paragraph{CRAG vs. NRAG}  
We refer to the full workflow described above—retrieval, fusion, and reranking with contextual chunking—as CRAG, short for Contextual Retrieval-Augmented Generation \cite{anthropic2024contextual}. For comparison, we also define a baseline, NRAG (Naive Retrieval-Augmented Generation), which uses basic document chunks in place of contextual ones.

\begin{figure}[t]
    \centering
    \includegraphics[width=0.7 \columnwidth, height=0.9\textheight, keepaspectratio]{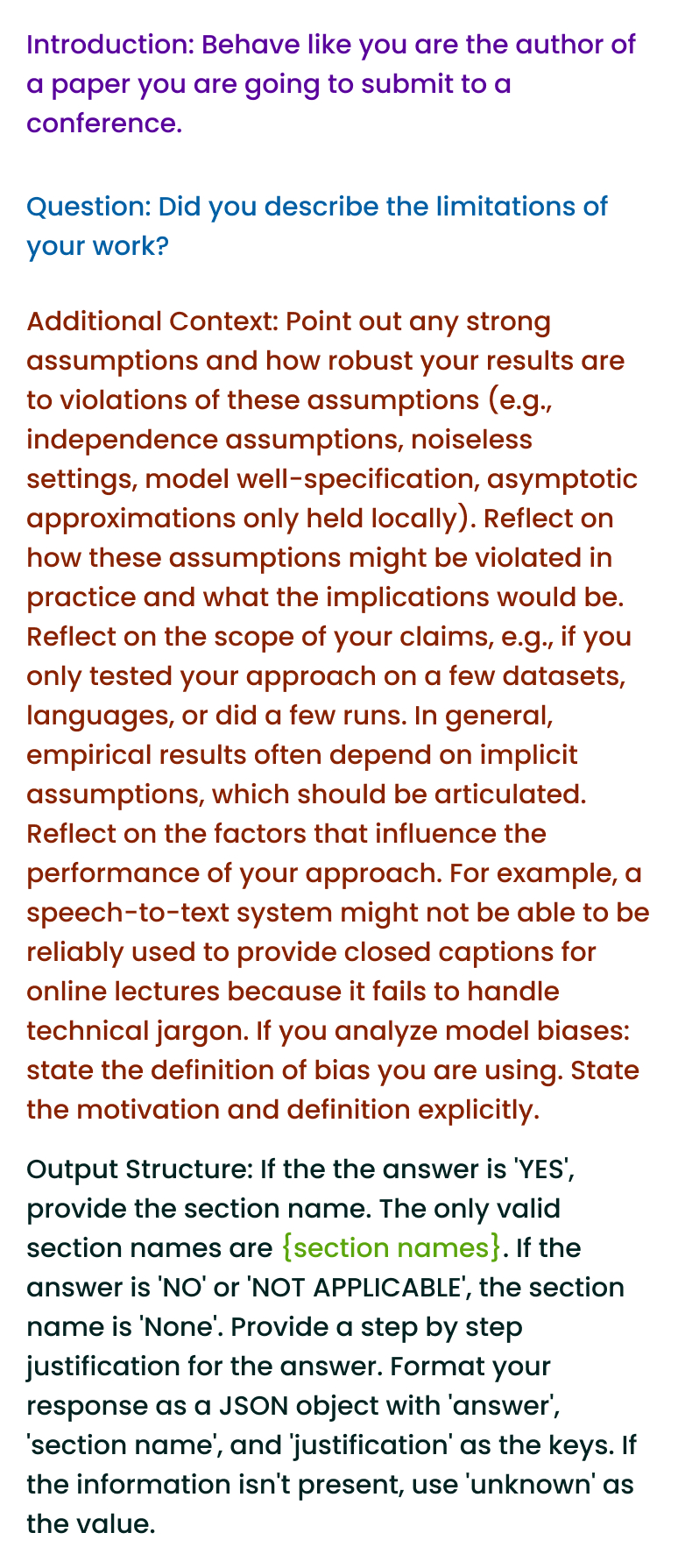}
    \caption{Example prompt for question A1. 
    \colorbox{Introduction!20}{\raisebox{0pt}[1.1ex][0.5ex]{\textit{Introduction}}}: 
    instructs the LM to assume the role of an author. 
    \colorbox{QuestionBlue!20}{\raisebox{0pt}[1.1ex][0.5ex]{\textit{Question}}}: 
    the checklist item the LM must address. 
    \colorbox{AdditionalContext!20}{\raisebox{0pt}[1.1ex][0.5ex]{\textit{Additional Context}}}: 
    provides supporting guidance drawn from checklist documentation. 
    \colorbox{OutputStructure!10}{\raisebox{0pt}[1.1ex][0.5ex]{\textit{Output Structure}}}: 
    instructs the LM to return JSON with fields for \texttt{answer}, \texttt{section name}, and \texttt{justification}. 
    \colorbox{SectionNames!20}{\raisebox{0pt}[1.1ex][0.5ex]{\textit{Section Names}}}: 
    lists valid section names parsed directly from the parsed files.}
    \label{fig:prompts}
\end{figure}

\subsection{Prompt Design}
\begin{figure*}[t]
     \centering
         \includegraphics[width=\textwidth]{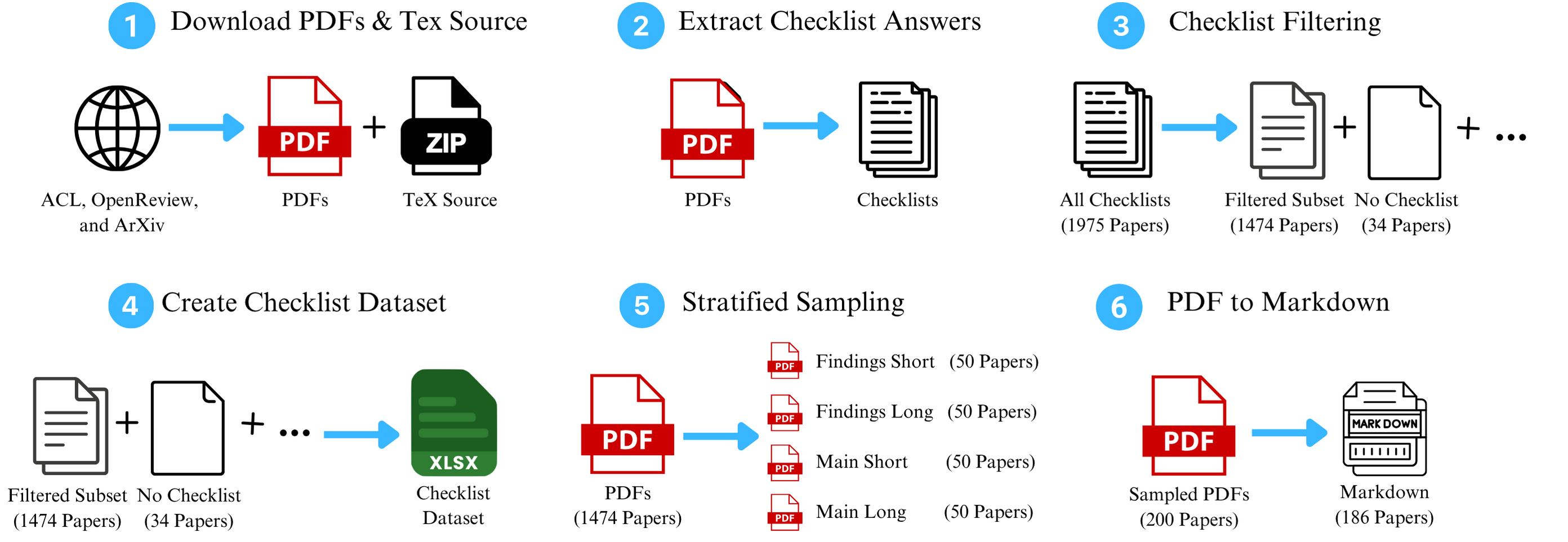}
    \caption{Checklist dataset generation pipeline with six stages: 
    (1) download PDFs and TeX sources for each paper, 
    (2) extract and verify checklist responses, 
    (3) analyze and filter out problematic checklists (e.g., blank or missing responses), 
    (4) organize the remaining checklists and metadata into a structured Excel file, 
    (5) apply stratified sampling across categories, and 
    (6) convert the sampled PDFs into Markdown using the \texttt{marker} library\footnote{\url{https://github.com/datalab-to/marker}} for evaluation.}
    \label{fig:ChecklistDataset}
 \end{figure*}

ConfReady uses modular instructions that can be adapted to different conference checklists and checklist versions. The ACL 2023 checklist, for instance, contained \fpeval{\numquestionsint} questions labeled A1–D5. 
Each question is mapped to a dedicated prompt with a uniform structure: Introduction, Question, Additional Context, and Output Structure. Figure~\ref{fig:prompts} shows the prompt for A1 (\textit{``Did you discuss the limitations of your work?''}).

The prompt is designed to provide the LM with the same information humans should consider when answering the question. The "Question" corresponds to an individual question in the checklist. The "Additional Context" is information provided from  Guidelines for Answering Checklist Questions on the ACL Rolling Review website\footnote{\url{https://aclrollingreview.org/responsibleNLPresearch/}}.

The \textit{Output Structure} specifies that the response should be a JSON object with \texttt{answer}, \texttt{section name}, and \texttt{justification} as keys. 
Restricting output to JSON has been shown to improve classification accuracy and reproducibility \cite{tam2024letspeakfreelystudy, es-etal-2024-ragas}. 
Section names are taken directly from the parsed files, so the LM is limited to choosing among sections that appear in the paper.

\subsection{Frontend with Generated Responses} 
After inference, the LM output is passed to the frontend in a structured JSON format with three fields: \texttt{answer}, \texttt{section name}, and \texttt{justification}. 
When the answer is \texttt{"Yes"}, the corresponding section name is displayed in the interface (Figure~\ref{fig:UIChecklist}). 
When the answer is \texttt{"No"}, the section name defaults to \texttt{"unknown"}, and the justification is shown instead.

\paragraph{Backend Hallucination Mitigation}
 
RAG applications are known to reduce hallucination inherent in LMs \cite{shuster-etal-2021-retrieval-augmentation}. To further mitigate the two hallucination categories, factuality and faithfulness, outlined by \citet{huang2023surveyhallucinationlargelanguage}, the application implements several safeguards. First, to address instruction inconsistency (a type of faithfulness hallucination) where the LM deviates from the instruction to return a single section, the system withholds a response. Second, because section names are parsed directly from the paper, only those names are allowed as valid answers. 

\paragraph{User Checklist Modification} LM-generated answers are intended to assist authors rather than replace their judgment. Users must review each response for accuracy, and questions concerning the use of AI assistants in research, coding, or writing must be answered manually. As a final safeguard, authors are required to validate responses before export. The validated checklist can then be exported as a Markdown document. Markdown was chosen because it is lightweight, easy to convert into other formats (e.g., PDF and TeX), and widely adopted in open-source ecosystems such as GitHub READMEs and Hugging Face model cards \cite{yang2024navigating}.

\subsection{System Architecture}
\paragraph{User Interface} The frontend is built with React\footnote{\url{https://reactjs.org}},  a JavaScript library for creating modular and interactive web applications. For consistent and responsive styling across devices, we use TailwindCSS\footnote{\url{https://tailwindcss.com/}}, a utility-first CSS framework that accelerates UI development with predefined class utilities.

\paragraph{API Orchestration and Backend Workflow}
The backend is implemented using Flask\footnote{\url{https://flask.palletsprojects.com/en/3.0.x/}}, a lightweight Python web framework that manages communication between the frontend and backend. It handles file uploads, runs processing scripts (e.g., TeX parsing), and orchestrates interactions with the RAG pipeline. RAG is implemented by LlamaIndex \citep{Liu_LlamaIndex_2022}, which integrates external context into the LM inference. To maintain real-time communication between the backend and frontend, a server-side event endpoint on the Flask server streams updates to the client during critical stages of the file processing workflow.

\section{ConfReady Dataset}
\label{sec:dataset}

To evaluate ConfReady and understand how authors engage with checklist questions in practice, we curate a dataset of \numpapers ACL 2023 papers, the most recent major Association for Computational Linguistics venue to publish checklists alongside accepted submissions. While prior work has documented issues with checklist reliability—such as identical answers across questions or unsupported claims of code availability \cite{magnusson-etal-2023-reproducibility}—no prior work has open-sourced a large dataset of real, author-written checklist responses. Our dataset addresses this gap, supporting both large-scale analysis of checklist quality and benchmarking of RAG and LM systems. 

Figure~\ref{fig:ChecklistDataset} illustrates the six-stage pipeline used to construct the dataset: 
(1) download PDFs and arXiv TeX sources for each paper, 
(2) extract and verify checklist responses, 
(3) analyze checklists and create a filtered subset without issues (e.g., blank, missing, etc.), 
(4) organize the remaining checklists and metadata into a structured Excel file, 
(5) apply stratified sampling to select 50 papers from each category (ACL Findings Short, Findings Long, Main Short, Main Long), and 
(6) convert the sampled PDFs into Markdown using the \texttt{marker} library\footnote{\url{https://github.com/datalab-to/marker}}, leaving \numpaperseval papers after excluding a small number with conversion failures. Due to this, ConfReady prefers TeX source input when available.

\begin{table}[ht]
\centering
\footnotesize
\renewcommand{\arraystretch}{1.1}
\setlength{\tabcolsep}{5.0pt} 
\begin{tabular}{lcccc}
    \toprule
    & \multicolumn{2}{c}{\textbf{Findings}} 
    & \multicolumn{2}{c}{\textbf{Main}} \\
    \cmidrule(lr){2-3} \cmidrule(lr){4-5}
    \textbf{Metric} & Short & Long & Short & Long \\
    \midrule
    \multicolumn{5}{@{}l}{\textbf{All Collected Papers}} \\
    Total Papers & 189 & 712 & 164 & 910 \\
    No Checklist & 6 & 11 & 3 & 14 \\
    Blank Checklist & 8 & 36 & 10 & 60 \\
    All Yes Responses & 5 & 7 & 2 & 9 \\
    No Section Names & 7 & 15 & 3 & 28 \\ 
    AI Use in Writing & 13 & 39 & 11 & 60 \\ 
    Not on arXiv & 50 & 190 & 37 & 190 \\
    \midrule
    \multicolumn{5}{@{}l}{\textbf{Evaluation Sample}} \\
    Total Papers & 46 & 43 & 47 & 50 \\
    Avg Tokens (Paper) & 12563 & 19865 & 14049 & 24547 \\
    \bottomrule
\end{tabular}
\caption{Summary statistics grouped by track (Findings/Main) and paper length (Short/Long).}
\label{tab:checklist_acl_only}
\end{table}

\paragraph{Dataset Statistics} 
To assess the quality and consistency of checklist submissions, we conducted a detailed analysis of all \numpapers papers. Table~\ref{tab:checklist_acl_only} summarizes key statistics, revealing several notable inconsistencies. Some papers omitted checklists entirely (\textit{No Checklist}), while others appended blank templates (\textit{Blank Checklist}) or didn't reference a single section in their responses (\textit{No Section Names}). Notably we also had relatively few disclosures about AI use in writing (\textit{AI Use in Writing}). Finally, to support evaluations of RAG
and standalone LM backends on TeX, we also filtered out papers without corresponding arXiv TeX sources.

\paragraph{Token Length Analysis}
Figure~\ref{fig:AverageTokens} presents token count distributions across the four evaluation subsets. On average, ACL Main papers are longer than ACL Findings papers in both short and long categories.

\paragraph{Final Structure} 
The ConfReady dataset preserves parsed checklist question–answer pairs, justification text, referenced sections, and metadata flags indicating issues such as incomplete or blank submissions. Further details on the data collection and analysis process are provided in Appendix~\ref{sec:dataset_details}.

\begin{figure}[t]
    \centering
    \includegraphics[width=\columnwidth]{figures/token_distributions_all.png}
    \caption{Token count distributions for ACL 2023 papers across four categories in our evaluation subset: \colorbox{FindingsShortColor!50}{\raisebox{0pt}[1.1ex][0.3ex]{ACL Findings Short}}, \colorbox{FindingsLongColor!50}{\raisebox{0pt}[1.1ex][0.3ex]{ACL Findings Long}}, \colorbox{MainShortColor!50}{\raisebox{0pt}[1.1ex][0.3ex]{ACL Main Short}}, and \colorbox{MainLongColor!50}{\raisebox{0pt}[1.1ex][0.3ex]{ACL Main Long}}. The y-axis indicates the percentage of papers; the x-axis shows token counts (measured using the Llama-3 tokenizer). Red dashed lines indicate mean token counts.}
    \label{fig:AverageTokens}
\end{figure}

\section{Evaluation}
\label{sec:evaluation}

\begin{table}[ht]
    \centering
    \footnotesize
    \renewcommand{\arraystretch}{1.1}
    \setlength{\tabcolsep}{4.8pt}
    \begin{tabular}{lcccc}
        \toprule
        & \multicolumn{1}{c}{} & \multicolumn{1}{c}{\textbf{Findings}} 
        & \multicolumn{1}{c}{} & \multicolumn{1}{c}{\textbf{Main}} \\
        \cmidrule(lr){3-3} \cmidrule(lr){5-5}
        \textbf{Model} 
        & & Long 
        & & Long \\
        \midrule
        \multicolumn{5}{l}{\textbf{RAG Framework on TeX}} \\
        CRAG (Llama-3.1-405B) 
        & & 81.72 
        & & 81.93 \\
        CRAG (Llama-3.3-70B) 
        & & 78.07 
        & & 78.78 \\
        CRAG (GPT-4o) 
        & & 80.58 
        & & 79.86 \\
        NRAG (Llama-3.1-405B) 
        & & 78.44 
        & & 77.36 \\
        NRAG (Llama-3.3-70B) 
        & & 74.64 
        & & 73.65 \\
        NRAG (GPT-4o) 
        & & 80.43 
        & & 73.22 \\
        \midrule
        \multicolumn{5}{l}{\textbf{LM on TeX}} \\
        Llama-3.1-405B 
        & & 78.87 
        & & 75.83 \\
        Llama-3.3-70B 
        & & 78.69 
        & & 79.20 \\
        GPT-4o 
        & & 80.54 
        & & 78.45 \\
        \midrule
        \multicolumn{5}{l}{\textbf{LM on MD}} \\
        Llama-3.1-405B 
        & & 79.86 
        & & 77.26 \\
        Llama-3.3-70B 
        & & 80.97 
        & & 81.09 \\
        GPT-4o 
        & & 82.27 
        & & 77.38 \\
        \bottomrule
    \end{tabular}
    \caption{Accuracy comparison of RAG, LMs on TeX, and LMs on PDFs for ACL Main (Long) and ACL Findings (Long) papers.}
    \label{tab:rag_framework_llm_acl_column}
\end{table}

We evaluate Llama-3.1–405B, Llama-3.3–70B \cite{grattafiori2024llama3herdmodels}, and GPT-4o \cite{gpt4} on the evaluation sample. Due to compute limits, experiments focus on long-form ACL submissions.  

Models are tested in three settings: (i) RAG on TeX with CRAG and NRAG, (ii) LM on parsed TeX, and (iii) LM on MD. Human-annotated answers serve as references, allowing us to evaluate how effectively models can reflect on ethical considerations, reproducibility, and societal impacts in each setup.
  
\paragraph{Results}
CRAG consistently outperforms NRAG, confirming the benefit of section-aware retrieval, RRF, and LM reranking (see Table~\ref{tab:rag_framework_llm_acl_column}). LM on Markdown performs competitively with LM on TeX, and in some cases better, echoing previous work showing that structured Markdown can improve model fidelity \cite{min-etal-2024-exploring, jain-etal-2025-autochunker, galarnyk-etal-2025-inclusively}. Nevertheless, ConfReady favors TeX input with RAG, since TeX parsing better preserves section structure and reduces extraction errors, whereas PDF-to-Markdown conversion is prone to failures that can disrupt retrieval and alignment. Finally, GPT-4o and other models often perform better on Findings than on Main papers, suggesting that longer Main submissions (see Figure~\ref{fig:AverageTokens}) introduce added difficulty for checklist answering.
   
\paragraph{Error Analysis}  
Figure~\ref{fig:LLMAccuracy} shows accuracy by checklist question for 
ACL Findings Long. 
Question C1—\textit{“Did you report the number of parameters in the models used, the total computational budget (e.g., GPU hours), and computing infrastructure used?”}- had the lowest LM accuracy. 
Appendix~\ref{sec:qualitative_analysis} presents typical failure cases on C1. 
A closer review of these disagreements revealed a recurring pattern: LMs often answered “NO” unless all three components were explicitly mentioned, while human justifications typically cited only section references (e.g., “Section 4” or “Appendix C1”) without clarifying what details were provided. This highlights a stricter model standard for completeness versus more lenient human interpretations.

\section{Conclusion and Future Work} This paper introduces ConfReady, a LM-based system which can be used to empower authors to reflect on their work and act as an assistant to help authors with conference checklists. With ConfReady, authors can get a LM checklist response that they use to reflect on their work or modify it before submitting. We hope that the open-source application will be responsibly used as an assistant and tool for reflection.
As a future work, we are still working on improving the project across several different directions:
\begin{itemize}
    \item \textbf{Local LMs}: While ConfReady currently uses commercial providers to avoid the overhead of self-hosting, future versions will support local open-weight models to enable private, offline usage in settings where data sensitivity or API constraints are a concern.
    \item \textbf{Other Conference Checklists}: ConfReady currently supports checklists from conferences under the Association for Computational Linguistics (e.g., ACL, COLING, CoNLL, EMNLP, and NAACL). While NeurIPS support has been implemented, adapting ConfReady to other venues will require adjustments for different checklist structures and question formats.
\end{itemize}
 
\section*{Ethics Statement}

\paragraph{Structured Output Format}
A major issue with incorporating LMs into applications is their failure to follow output format inconsistency (faithfulness hallucination). We mitigate this by requiring responses in JSON format, similar to the JSON mode in the OpenAI and Gemini APIs \citep{geminiteam2024geminifamilyhighlycapable}. Additionally, some libraries such as Instructor\footnote{\url{https://github.com/instructor-ai/instructor}}
 also require JSON.

\paragraph{User Reliance and Scope}  
Analysis of ChatGPT usage shows that non-work messages now comprise over 70\% of interactions, up from 53\% in 2024 \citep{chatterji2025chatgpt}. This reflects a broadening role for LMs in everyday life, including learning and creative expression, beyond their original productivity-focused scope. At the same time, models can provide fluent but incomplete or outdated answers when knowledge falls outside their training data \cite{shah2025reportedcutofflargelanguage}. These patterns highlight the importance of using ConfReady as an aid for reflection and editing, not as a replacement for author responsibility.

\bibliography{anthology,custom}
\bibliographystyle{acl_natbib}

\appendix

\section{Features of the User-Interface}
\label{app:features_userinference}
 
The features incorporated into the user interface, along with the underlying rationale for their design, are detailed below. These design decisions aim to enhance usability, ensure accessibility, and support an intuitive and efficient user experience throughout the checklist completion workflow.

\begin{enumerate}
    \item \textit{Side Bar/Upload:} The side bar incorporates the visual identity of the platform. It has been visualized to resemble file tabs to help users connect with the overarching action being performed using visual connotation \cite{davis2017visual}. It contains the upload function which allows users to upload their paper's TeX source (single .tex file or zipped folder) and visual indication of the user’s current action. The bottom of the sidebar contains a model selector and links to the documentation\footnote{\url{https://confready-docs.vercel.app/docs/walkthrough}} and GitHub\footnote{\url{https://github.com/gtfintechlab/ConfReady}} placed according to information hierarchy principles \cite{guizani2022decade}. 
    \item \textit{Conference Selection:} Users select a conference checklist. Currently, the platform allows users to select from ACL checklists, NeurIPS, and NeurIPS Datasets and Benchmarks (NeurIPS D\&B).
    \item \textit{Primary navigation:} The top bar of the interface provides the users with functionality of switching between sections. It also indicates the progress for each section, keeping the users informed through visually represented data \cite{nontasil2024investigating}.
    \item \textit{Secondary navigation:} To refrain from disrupting the user’s workflow while performing important tasks like checking or editing responses, the secondary navigation allows movement to the next page without needing to return to the primary navigation. The intention with this navigation is reducing extraneous cognitive overload.
    \item \textit{Response sections:} Responses are filled in and users need to verify responses.
    \item \textit{Download:} Only after users have reviewed each section are they allowed to download all of their responses from either the sidebar or from the Download button in the final section. The intention is to encourage users to be responsible for verification of AI-driven results \cite{afroogh2024trust}. 
\end{enumerate}

The ConfReady user journey is shown in the ConfReady documentation\footnote{\url{https://confready-docs.vercel.app/docs/walkthrough}}. To use the application, users upload the TeX source (single .tex file or zipped folder) or PDF. Next, in order to enhance task completion \cite{conrad2010progress}, a progress screen appears to let users know of the backend RAG progress. This feature was added to the platform after informal interviews where it was noted that users wanted to get some indication on how long they needed to wait before they can check/edit responses, and download results.

\section{Qualitative Analysis of Checklist Responses}
\label{sec:qualitative_analysis}

We provide qualitative insights into system behavior on ACL 2023 checklists. 
Details of dataset construction and statistics are provided in Appendix~\ref{sec:dataset_details}.

\subsection{Discrepancies on Question C1}
Question C1 of the ACL checklist asks: 
\textit{“Did you report the number of parameters in the models used, the total computational budget (e.g., GPU hours), and computing infrastructure used?”} 
To better understand the low accuracy on this question, we reviewed a sample of cases where the LM response diverged from the human-provided checklist. 
Table~\ref{tb:c1_examples} shows representative examples. 
In all cases, human annotators answered "YES" while the LM answered "NO." 
This reflects a recurring failure mode: the LM applied a stricter standard, requiring explicit mentions of model parameters, compute budget, and infrastructure to justify a "YES," while human authors often responded affirmatively based on partial evidence.  

Notably, compute resource reporting was almost always absent. 
The LM typically acknowledged some modeling information but still answered "NO" if all three elements were not present. 
This conservative behavior highlights a mismatch between human and model expectations of completeness. 

\begin{table*}[h!]
\centering
\renewcommand{\arraystretch}{1.5}
\small
\begin{tabular}{p{2.5cm} p{2.5cm} p{3cm} p{6cm}} 
\toprule
\textbf{Human Response} & \textbf{LM Response} & \textbf{Human Justification} & \textbf{LM Justification} \\
\midrule
 Yes & No & Section A.1 & The text does not discuss the experimental setup, including hyperparameter search and best-found hyperparameter values. While it mentions experiments and evaluations of different models, it does not provide details on how hyperparameters were tuned or selected. \\
 
Yes & No & Appendix D & The paper does not report the number of parameters in the models used, the total computational budget, or computing infrastructure used. Although it mentions the use of specific models like BERT and the number of GPUs used for training, it does not provide detailed information about the model sizes or computational resources.\\

Yes & No & Section 3.2 & The text does not mention the number of parameters in the models used, the total computational budget (e.g., GPU hours), or computing infrastructure used. The models used are RoBERTa-large models, but the text does not provide information about the number of parameters or computational resources used. \\

Yes & No &  Please see Section 4.1. & The text does not mention the number of parameters in the models used, the total computational budget (e.g., GPU hours), or computing infrastructure used.  \\

\bottomrule
\end{tabular}
\caption{
Discrepancies on checklist question C1 in ACL Findings Long. The justifications provided are from CRAG and MD outputs generated using Llama 3.1 405B. All example have human answers ``Yes'' and LM answers ``No,'' reflecting stricter model standards for reporting parameters, compute budget, and infrastructure. 
}
\label{tb:c1_examples}
\end{table*}

\subsection{ACL Edge Cases} 
While the ConfReady system accurately processed most ACL submissions, several edge cases highlighted current limitations. 
One issue involved exceeding the model's context length limit, leading to inference failures. 
Additionally, some model-generated responses were phrased awkwardly or contained placeholder-like text (e.g., "x and y should be..."), suggesting room for prompt refinement or output filtering.

\section{Checklist Dataset Construction}
\label{sec:dataset_details}

We describe the pipeline used to construct the structured checklist dataset.

\subsection{Data Collection and Extraction}

Due to the scale of the dataset, most of the collection was automated using Python scripts. 
The multi-stage pipeline proceeded as follows:

\begin{enumerate}
    \item \textbf{Link Retrieval} Automated scripts using \texttt{requests} and \texttt{BeautifulSoup} were used to identify and retrieve arXiv links as well as ACL Anthology PDF links for each paper. 
    \item \textbf{PDF Retrieval and Preprocessing} Normalized PDF links were obtained from ACL Anthology URLs. To extract checklists, only the last two pages of each ACL paper were parsed using \texttt{PyPDF2}, since checklists consistently appeared there.
    \item \textbf{Checklist Parsing} Each checklist was parsed using conference-specific regex templates. ACL checklists featured labeled questions (e.g., A1) with symbolic ticks and optional justifications; answers were matched using proximity-based heuristics. 
    \item \textbf{Structured Storage} The extracted checklist information was standardized, manually reviewed, and stored in structured Excel sheets for incremental updates. Each paper was also mapped to its arXiv \texttt{.tar.gz} TeX source to allow linking between TeX and PDF files. 

\end{enumerate}

\begin{figure*}[t]
    \centering
    \includegraphics[width=\textwidth, keepaspectratio]{figures/llm_accuracy_plot.png}
\caption{
Accuracy by checklist question for 
\colorbox{FindingsLongColor!50}{\raisebox{0pt}[1.1ex][0.3ex]{ACL Findings Long}} on Llama-3.1-405B across four setups: CRAG, NRAG, LM on TeX, and LM on MD. 
}

    \label{fig:LLMAccuracy}
\end{figure*}
\label{app:}
\end{document}